\documentclass[pmlr,twoside,11pt]{article}
\usepackage{jmlr2e}

\usepackage{times}
\usepackage{float}
\usepackage{epsfig}
\usepackage{graphicx}
\usepackage{amsmath}
\usepackage{amssymb}
\usepackage{multirow}
\usepackage{algorithm}
\usepackage{subcaption}
\usepackage{algpseudocode}
\usepackage{scrextend}
\usepackage{tabularx}
\usepackage{booktabs}
\usepackage{bbm}
\usepackage{breqn}
\usepackage{verbatim}
\usepackage{wrapfig}
\usepackage{multicol}
\usepackage{geometry}
\usepackage{changepage}
\usepackage{textcomp}

\newcommand{\vect}[1]{\boldsymbol{#1}}

\jmlrheading{66}{2017}{1-16}{5/17}{08/17}{Dianbo Liu , Fengjiao Peng, Andrew Shea, Ognjen (Oggi) Rudovic, Rosalind Picard}

\ShortHeadings{DeepFaceLIFT}{Liu et al.}
\firstpageno{1}
\begin{document}

\title{DeepFaceLIFT: Interpretable Personalized Models \\ for Automatic Estimation of Self-Reported Pain}

\author{\name Dianbo Liu*$^{2,3}$ \email dianbo@mit.edu \\
\name Fengjiao Peng*$^{1}$ \email fpeng@mit.edu \\
\name Andrew Shea*$^{3}$ \email ashea@mit.edu \\
\name Ognjen (Oggi) Rudovic$^{1}$ \email orudovic@mit.edu \\
\name Rosalind Picard$^{1}$ \email picard@media.mit.edu \\\\
       \addr $^1$Media Lab, MIT, Cambridge, MA, USA
       \AND
       \addr $^2$Computer Science and Artificial Intelligence Laboratory, MIT, Cambridge, MA, USA
       \AND
       \addr $^3$Department of Electrical Engineering and Computer Science, MIT, Cambridge, MA, USA}


\maketitle

\begin{abstract}

Previous research on automatic pain estimation from facial expressions has focused primarily on ``one-size-fits-all"
metrics (such as PSPI). In this work, we focus on directly estimating each individual's self-reported visual-analog scale (VAS) pain metric, as this is considered the gold standard for pain measurement. The VAS pain score is highly subjective and context-dependent, and its range can vary significantly among different persons. To tackle these issues, we propose a novel two-stage personalized model, named DeepFaceLIFT, for automatic estimation of VAS. This model is based on (1) Neural Network and (2) Gaussian process regression models, and is used to personalize the estimation of self-reported pain via a set of hand-crafted personal features and multi-task learning. We show on the benchmark dataset for pain analysis (The UNBC-McMaster Shoulder Pain Expression Archive) that the proposed personalized model largely outperforms the traditional, unpersonalized models: the intra-class correlation improves from a baseline performance of 19\% to a personalized performance of 35\% while also providing confidence in the model\textquotesingle s estimates -- in contrast to existing models for the target task. Additionally, DeepFaceLIFT automatically discovers the pain-relevant facial regions for each person, allowing for an easy interpretation of the pain-related facial cues.
\end{abstract}
\let\thefootnote\relax\footnotetext{*These authors contributed equally to the manuscript.}

\section{Introduction}

Research on automatic estimation of pain levels from facial expressions has focused primarily on estimating (arguably) objective metrics rather than self-reports. These objective metrics, such as the Prkachin and Solomon Pain Intensity (PSPI) \citep{Prkachin2008}, use facial action intensities to quantify facial expressions and score pain. Using these metrics, two individuals who make the same facial expression should be given very similar pain scores. However, because pain is a fundamentally subjective experience \citep{textbookPain21} and because individuals express pain differently depending on qualitative factors such as their motor ability, gender, and age \citep{fillingim2009sex,kunz2008impact,scipio2011perception}, these objective metrics may not sufficiently capture an individual's true pain. Furthermore, many studies have found low correlations between facial expressions and self-reported pain levels \citep{Prkachin2008}. These observations pose the question of how well automatic estimates of objective metrics capture an individual's actual experience. Self-reported pain scores are often considered the \textit{gold standard} because they provide information about an individual's subjective experience \citep{craig1992facial}. Therefore, an approach that automatically estimates self-reported pain scores rather than objective pain metrics may be more useful for individuals experiencing pain.

\begin{figure}[h]
	\centering
	\includegraphics[width=0.6\textwidth]{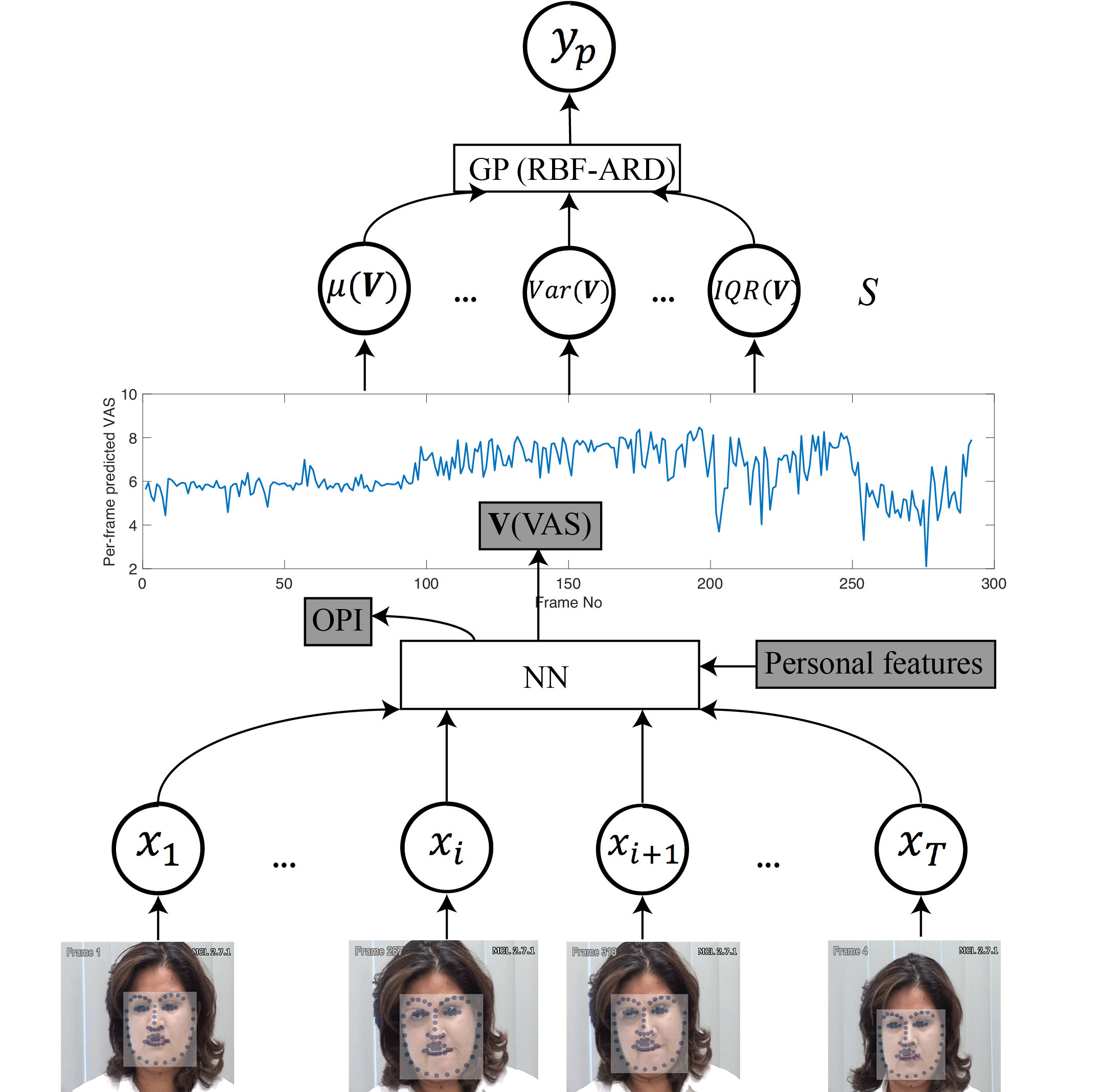}
 	\caption{Graphical representation of the proposed DeepFaceLIFT two-stage learning model for using AAM facial landmarks ($x$) extracted from each frame of an image sequence to estimate sequence-level VAS ($y_p$). The NN outputs frame-level estimates of VAS that are concatenated to form $\mathbf{V}$, which is then used to compute various statistics that form the vector $\mathbf{S}$. The statistics vector $\mathbf{S}$ is then passed to a Gaussian process regression model to estimate sequence-level VAS.}
    \label{fig:twostage}
\end{figure} 
Because individuals make different facial expressions depending on personal factors, a method for automatic pain estimation would likely improve by not only accounting for how facial expressions and specific facial actions vary, but also personalizing on an individual's qualitative characteristics by explicitly representing them as features. These features could then be incorporated into a model. This feature-level personalization differs from the model-level personalization (i.e., varying the model architecture) as a single model is consistently applied to all subjects. Using these personalized features may enable the model to learn additional personal information and increase its estimative power.

Another important aspect of personalized applications is ensuring that model outputs can be traced back to specific components of the input features, as doing so is crucial for explaining the specific characteristic and behavioral differences between individuals. This problem is broadly faced in deep learning research due to the lack of interpretability of learned representations with respect to the original input features \citep{zeiler2014visualizing,shrikumar2016not}. For example, some methods have shown that personalizing models improve outputs \citep{zen2014unsupervised,clifton2013gaussian}; however, the model outputs and learned internal representations cannot be easily nor intuitively linked to specific portions of the original inputs. As a result, while the improved performance achieved through personalization indicates that qualitative differences indeed exist between training subjects, exactly what these differences are and how they affect the final output are not clear. Therefore, having the ability to examine how specific aspects of input features contribute to final outputs would provide insight into what differences exist among subjects and how these differences influence model estimations.

To address these challenges of subjectivity, personalization, and interpretability, we propose a two-stage hierarchical learning algorithm called DeepFaceLIFT (LIFT = Learning Important Features) \citep{shrikumar2016not} that estimates pain scores obtained using the self-reported visual-analog scale (VAS) \citep{carlsson1983assessment}. The first stage of the model comprises a fully-connected neural network (NN) that takes raw Active Appearance Model (AAM) facial landmark coordinates \citep{unbc} as input features. The second stage then processes the NN outputs and passes them as features into a Gaussian process (GP) regression model. In addition to producing accurate VAS estimates, the model also provides an interpretable learning framework. This is accomplished in the first stage by leveraging an existing model called DeepLIFT \citep{shrikumar2016not} to examine inputs and in the second stage by utilizing a Radial Basis Function kernel with Automatic Relevance Determination (RBF-ARD)  to examine the learned GP kernel weights. We show on the UNBC-McMaster Shoulder Pain Expression Archive Database \citep{unbc} that the proposed personalized approach for automatic VAS estimation outperforms traditional approaches. The model outline is shown in Fig.\ref{fig:twostage}.


\section{Related Work}

Although much research exists on automated recognition of affect from human facial expressions (for surveys, see \citet{Zeng2009,tac_ref58}),  until recently, only a handful of works have focused on  automated pain estimation. Due to advances in computer vision and the recent release of the UNBC-McMaster dataset \citep{unbc}, pain analysis from face images has seen significant advances. This dataset provides videos where each frame has been rated using the PSPI score. Although VAS is still the most commonly accepted pain score in clinical settings, all existing automatic methods for pain estimation from pain images have focused on estimating PSPI scores. We outline below the recently published works.


Face shape features have been used in NNs to classify images of faces showing typical mood versus pain expressions. \citet{monwar2006pain}, \citet{Lucey2011} and \citet{Ashraf2009} used AAM-based features combined with Support Vector Machine (SVM) classifiers to classify pain versus no-pain images. Rather than treating pain as binary classification task, \citet{Hammal2012} attempted to estimate pain intensity on a 4-level scale using one-versus-all SVM classifiers, and \citet{Kaltwang2012} performed estimation of the full 15 level PSPI scale using Relevance Vector Regression models. Likewise, \citet{rudovic2013automatic} proposed a Conditional Random Field (CRF) model with heteroscedastic variance, showing that changing the model variance provides a better fit for individual pain sequences. More recently, several works using deep learning have been proposed for PSPI estimation \citep{egede2017fusing,Rodriguez2017}.
 
Other works have attempted to estimate the pain intensity of entire sequences. \citet{Rusk2016} proposed a Hidden Conditional Random Field (HCRF) and semi-supervised learning method to identify the peak of pain in image sequences. \citet{Sikka2013} proposed a multi-instance learning method to identify the most expressive (in terms of pain) segments within image sequences. However, these sequence-level labels are derived using heuristics and do not relate to established pain ratings such as VAS. More recently, \citet{martinez2017personalized} proposed a personalized model for VAS estimation based on a combination of Long-short Term Memory Networks (LSTMs) \citep{mikolov2010recurrent} and HCRFs, where the parameters of the latter are modulated by the person's 'expressiveness score' - a heuristic derived from previous VAS rating by that person. While this method showed promising results in improving the VAS estimation performance compared to traditional un-personalized models, it does not generalize to previously unseen subjects until a certain number of their ratings has been collected. By contrast, the approach proposed here is designed to circumvent this limitation by generalizing (in a personalized manner) to new persons without the explicit need for their VAS data prior to applying the model.

Aside from the UNBC-McMaster dataset \citep{unbc}, there have been other pain recognition studies based on other datasets (e.g., see \citet{werner2012pain,tac_ref66}). Pain detection has been attempted using physiological signals such as brain hemodynamic responses using NIRS \citep{Aasted2016,Yucel2015}. Nevertheless, these works have neither attempted automatic estimation of VAS nor in a personalized manner.

\section{DeepFaceLIFT: The Method}
\label{pmodel}
We used the UNBC-McMaster Shoulder Pain Expression Archive Database (UNBC-PAIN) \citep{unbc}, a publicly available dataset containing facial videos (we refer to these as image sequences for clarity) of subjects who suffer from one-sided shoulder pain as they undergo various range-of-motion exercises. The dataset contains 200 image sequences gathered from 25 subjects, totaling 48,398 image frames. Each frame has a PSPI score obtained on a 15-point ordinal scale using facial action unit (AU) intensities. The PSPI rating is computed as a sum of specific AU intensities as:
\begin{equation}
PSPI = AU4 + max(AU6,AU7) + max(AU9,AU10) + AU43
\end{equation}
For each image sequence, the dataset also includes observed pain index (OPI) scores that were rated by experienced pain-observers on a 0-5 Likert-like scale \citep{unbc}. The dataset also includes for each image sequence the self-reported VAS rating, which is defined on an 0-10 ordinal scale. When estimating subjects' VAS scores, we do not use PSPI during training nor testing. We do, however, use the OPI scores in certain settings. Although the UNBC-PAIN dataset contains a large amount of image frames (48,398), estimating VAS scores poses a challenge due to the comparatively limited number of sequence-level VAS labels (200).

Additionally, we obtain three manually-labeled features to personalize the method. These features are derived from each subject's appearance and are complexion, age, and gender. Complexion is divided into three bins using a Fitzpatrick-like skin-tone scale: pale-fair, fair-olive, and olive-dark \citep{fitzpatrick1988validity}. Age is also divided into three bins: young, middle-aged, and elderly. Finally, gender is divided into two bins: male and female. These bins were independently assigned then agreed upon by the primary authors. These personal features were selected based on their straightforward labeling as well as their relevance to how individuals express pain. For example, individuals of different age and gender have been shown to express pain differently. Complexion is used because it is easily observed and thought to loosely contain information about an individual’s medical information, background, and/or daily habits. 

For input features, we use AAM facial landmarks instead of raw images. We do so primarily for two reasons. The first is to reduce the feature space of each feature from 320x240 (image pixels) to 132 (66 concatenated (x, y) coordinates). The second and main reason is to ensure that the feature representations are easily distinguished and intuitive. Furthermore, doing so allows us to apply DeepLIFT (see Sec. 3) to interpret how individual facial landmarks vary across persons and contribute to the output VAS scores. 
  
{\bf Notation.} The estimation of VAS is formulated as a regression problem, where we are given $N_i$ image sequences of person $i\in \{0,\dots,P\}$ where \textit{P} is the number of target persons. The sequences of each person are annotated in terms of VAS as $\mathcal{V}=\{V_{1},\dots,V_{P}\}$, where $V_{i}=\{v_{i}^{1},\dots,v_{i}^{N_i}\}$ and individual per-sequence VAS scores are $v_i\in \{0,...,10\}$. For OPI scores we have: $\mathcal{O}=\{O_{1},\dots,O_{P}\}$, where $O_{i}=\{o_{i}^{1},\dots,o_{i}^{N_i}\}$ and $o_i\in \{0,...,5\}$. Additionally, let $f \in \{1, \dots, \mathcal{F}\}$ denote the frame number and $s \in \{1, \dots, \mathcal{S}\}$ denote the sequence number. Each sequence is represented by a set of input features $X_s=\{x_1, \dots,x_{L_s}\}$, where $L_s$ is duration or the number of frames in a given sequence $s$.

\subsection{Stage 1: Weakly Supervised Neural Network}
\label{stageone}
To leverage all frame-level data before making sequence-level estimations, the first stage of DeepFaceLIFT applies multi-instance and multi-task learning to train a fully-connected NN 
(Fig.\ref{fig:twostage}). This network uses frame-level AAM facial landmark features as input, where each feature is labeled with the VAS score of its corresponding sequence. Applying a NN is appropriate in this setting as the facial landmarks may be related in highly nonlinear manners, and these relations can be learned automatically using NNs. DeepFaceLIFT uses a 4-layer NN with ReLU activation functions in the first stage of training to generate frame-level VAS scores. The same hidden layer architecture is used in all experiments, where each hidden layer in the NN contains 300, 100, 10 and 100 nodes, respectively. The NN is trained for 100 epochs with a batch size of 300. 

In the first stage, DeepFaceLIFT generates frame-level VAS scores in 3 separate settings to compare different personalization strategies. Each setting differs in how personal information is used in the model (Not used, Appended to the 3rd layer of NN, Appended to Input Features). Table 1 shows these settings in the first column labeled with S1. Within each of these settings, another 2 separate labeling schemes are used to train the NN in a multi-task learning manner (Labeled with VAS, Labeled with both VAS and OPI).\footnote{For more details on multi-task learning and recent advances, please see Caruana's introduction and Ruder's overview \citep{caruana1998multitask,ruder2017overview}} Table 1 shows these settings in the second column labeled with S1. Thus, for each frame $f \in \{1, \dots, M\}$, where M is the total number of frames in the dataset, the NN estimates either $V_f$ or both $V_f$ and $O_f$.

\subsection{Stage 2: GP Model with RBF-ARD Kernel}
\label{stagetwo}
Various methods have been reported that capture temporal information in a sequence of image frames, such as Recurrent Neural Networks (RNNs) \citep{mikolov2010recurrent} and CRFs \citep{rudovic2013automatic}. Instead of using a temporal model, in the second stage of our personalized modeling, we consider the problem as a static one by extracting sequence-level statistics from the outputs of the first stage, then passing these statistics into the GP regression model in the second stage. Doing so enables the model to capture and encode both frame-level and sequence-level dependencies. More specifically, using the frame-level VAS estimations, DeepFaceLIFT computes sequence-level statistics (as described below) that are then fed as inputs into the GP model to obtain a final VAS estimate. The GP model is tested in two different settings. In the first setting, the input statistics are calculated using only the frame-level VAS estimates. In the second setting, both frame-level VAS and OPI estimates are used. Table 1 shows these settings the first column labeled with S2. 

Because the image sequences vary in length, the model creates a consistently sized feature vector by computing a set of statistics for each sequence $s$. These statistics form the vector $\vect{S_s}=$ \{$\mu(V)=$ mean, $\eta(V)=$ median, $Min(V)=$ minimum, $Max(V)=$ maximum, $Var(V)=$ variance, $\mu^3(V)=$ 3rd moment, $\mu^4(V)=$ 4th moment, $\mu^5(V)=$ 5th moment, $Sum(V)=$ sum of all values, $IQR(V)=$ interquartile range\}. These \textit{sufficient statistics} ~\citep{hogg1995introduction} are chosen because they  capture diverse and important information that may be used to infer unknown parameters in many distribution families~\citep{hogg1995introduction}. When using only the frame-level VAS estimates to compute these statistics, the length of $\vect{S_s}$ is 10; when using both the frame-level VAS and OPI estimates, the length of $\vect{S_s}$ is 20, as two  sets of statistics are calculated. $\vect{S_s}$ is then fed as input to the GP regression model \citep{rasmussen2006gaussian}. Formally, the GP regression model is formulated as follows: 
\begin{equation}
 \vect{y} = f(\vect{S}) + \epsilon
\end{equation}
where $\epsilon \sim \mathcal{N}(0, \sigma^2_{v})$ is i.i.d. additive Gaussian noise. The objective is to infer the latent functions $f$ given the training dataset $\mathcal{D} = \{ \vect{S}, \vect{Y} \}$. Following the GP framework, we place a prior on the
function $f$ so that the function values $\vect{f} = f(\vect{S})$ follow a Gaussian distribution $p(\vect{F}|\vect{S}) = \mathcal{N}(\vect{F}|\vect{0}, \vect{K})$. Here, $\vect{F} = \{\vect{f}\}$, and $\vect{K} = k(\vect{S},\vect{S}^T)$ is the kernel covariance function. Training the GP consists of finding the hyperparameters that maximize the log-marginal likelihood
{\small
\begin{align}
\small
\nonumber\log p(\vect{Y}|\vect{X}, \vect{\theta}) =&-\textrm{tr}\left[(\vect{K} + \sigma^2\vect{I})^{-1}\vect{Y}{\vect{Y}}^T\right] \\
\label{marg}  &- \log\vert \vect{K} + \sigma_v^2\vect{I} \vert + \textrm{const}.
\end{align}
}

\begin{figure}
	\centering
	\includegraphics[width=0.8\textwidth]{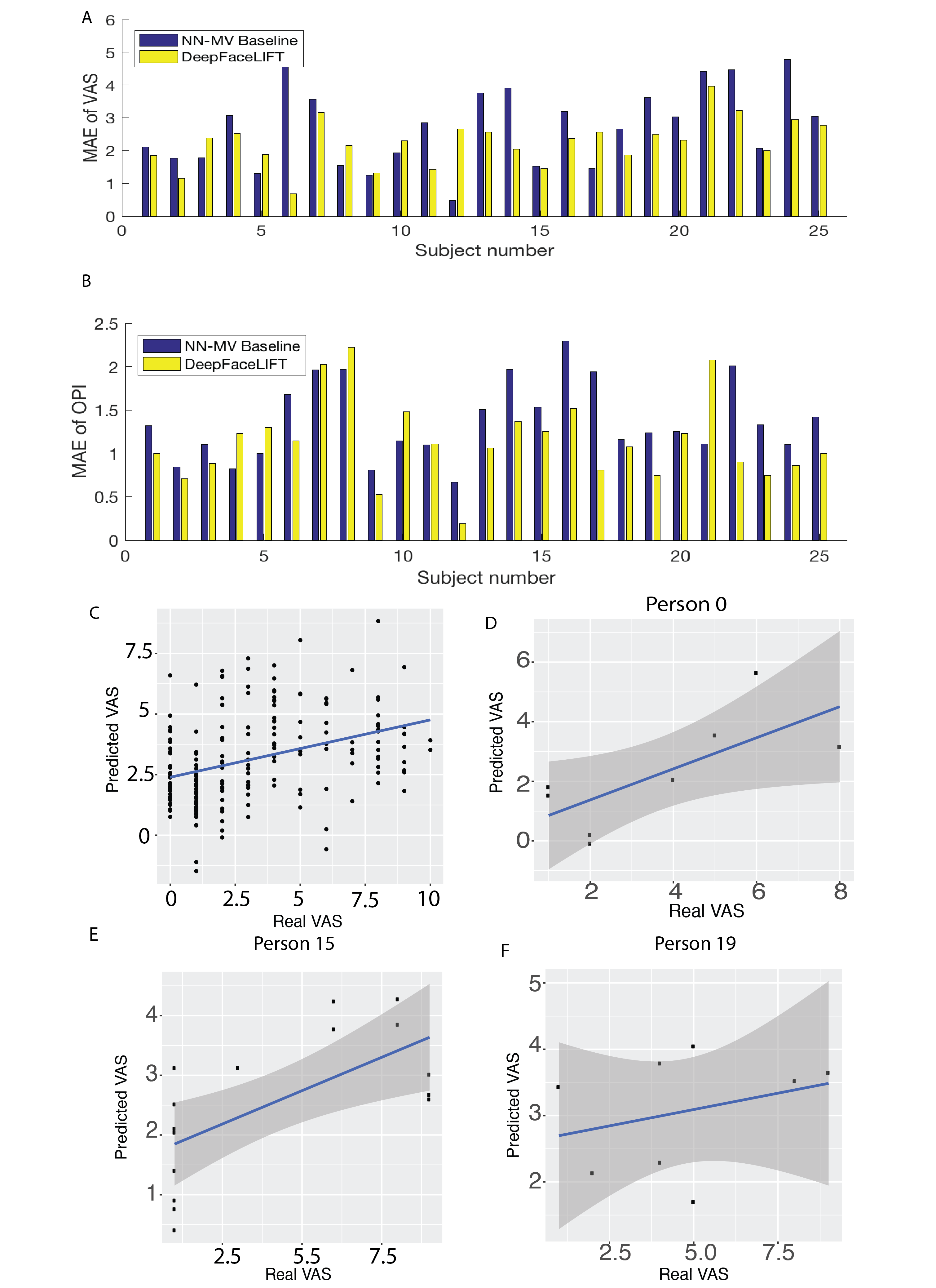}
    \caption{ DeepFaceLIFT results for estimation of VAS from AAM facial landmarks. A and B show the final Mean Absolute Error (MAE) scores for each subject when estimating A) VAS (average MAE=2.18) and B) OPI (average MAE=1.12). The final MAEs for 18 out of 25 subjects are lower than the NN-MV baseline. C-F show the estimated VAS versus the reported VAS along with the GP model's uncertainty estimate (shaded area) for four randomly chosen subjects. The GP model learns different uncertainty levels for different individuals; this may be attributed in part to input-feature noise, subjectivity in self-reported VAS scores, etc.}
    \label{fig:Results}
\end{figure}

Given a test input $\vect{s}_\ast$ we obtain the GP estimative distribution by conditioning on the training data $\mathcal{D}$ as $p(\vect{f}_\ast |\vect{s}_\ast, \mathcal{D} = \mathcal{N}(\mu(\vect{s}_\ast, V(\vect{s}_\ast)))$ with the mean and variance given by
\begin{align}
\small
\label{post_mu}\mu(\vect{s}_\ast) &= {\vect{k}_\ast}^T (\vect{K} + \sigma_v^2\vect{I})^{-1}\vect{Y}\\ 
\label{post_s}V(\vect{s}_\ast) &= k_{\ast\ast} - 
{\vect{k}_\ast}^T (\vect{K} + \sigma_v^2\vect{I})^{-1}
\vect{k}_\ast,
\end{align}
where {\small $\vect{k}_\ast = k(\vect{S}, \vect{s}_\ast)$} and {\small $k_{\ast\ast} = k(\vect{s}_\ast, \vect{s}_\ast)$}. For convenience, we denote 
$\vect{\mu}_\ast^{(v)} = \mu(\vect{s}_\ast)$ and $V^{(v)}_{\ast\ast} = V(\vect{s}_\ast)$. The RBF-ARD kernel \citep{rasmussen2006gaussian} is defined as
\begin{equation}
k(\vect{s},\vect{s}^\prime) = \sigma_f^2\exp(-(\vect{s} - \vect{s}^\prime)M(\vect{s} - \vect{s}^\prime)),
\end{equation}
where $M=diag(\ell_1,...,\ell_D)^{-2}$ is a diagonal matrix where each element indicates the importance of its corresponding statistic feature ($1,...,D$). Each value $\ell$ is an inverse of length-scale that indicates how relevant an input is to the output: at the extremes, if the length-scale is very large, then the covariance will be very low, indicating that the output is uncorrelated with the input and effectively removing that feature from the inference. Contrastingly, if the length scale is very small, then the opposite is true, indicating that the feature is of high relevance to the inference \citep{rasmussen2006gaussian}. The regression mapping is then fully defined by the set of hyperparameters $\vect{\theta} = \{\ell_1,...,\ell_D, \sigma_f, \sigma_v\}$. 

The application of this GP model is appropriate given the small number of input sequences (200), as nonparametric probabilistic GP models have the ability to generalize well on small datasets \citep{rasmussen2006gaussian}. The use of the RBF-ARD kernel is as well since the kernel provides uncertainties for its estimates, which is important for assessing the reliability of the estimated VAS.

To summarize, the learning and inference in this two-stage approach proceeds as follows: in the first stage, DeepFaceLIFT trains a NN using multi-task learning under different personalization strategies (Sec.\ref{stageone}). Next, the output of the trained NN is used to compute sequence-level statistics. These statistics are then passed into a GP model to obtain a personalized estimate of VAS (Sec.\ref{stagetwo}). Inference is done in a straightforward manner by propagating features through the two-stage model, resulting in the VAS and uncertainty estimate for each subject (see Fig.\ref{fig:Results}). The per-subject estimation results of VAS and OPI are also shown in Fig.\ref{fig:Results}.


\begin{table*}
\centering
\small
\caption{The performance of different methods tested for VAS (0-10) estimation in terms of MAE (ICC). Below are the results of the baseline models using NN-MV (Mean Voting), RNN, and HCRF. The model was also tested by appending personal features (PFs) to the GP inputs; however, doing so did not improve estimates.}
\label{my-label}
\begin{adjustwidth}{}{}
\begin{tabular}{|c|c|c|c||c|c|c|c|}
\hline
\multirow{2}{*}{\textbf{\scriptsize{S1: PFs}}} & \multirow{2}{*}{\textbf{\scriptsize{S1: NN Labels}}} & \multirow{2}{*}{\textbf{\scriptsize{S2: GP Input}}} & \multirow{2}{*}{\textbf{\scriptsize{S2: Output}}}  &\multicolumn{3}{|c|}{\textbf{Baselines Results}} \\ \cline{5-7}
    							          &										                   &                                              &                   & NN-MV & RNN & HCRF           \\ \hline
\multirow{3}{*}{None }   & \multirow{2}{*}{\textbf{VAS,OPI}}     & VAS,OPI 			          & 2.32 (0.25)         	  &\multirow{2}{*}{2.56 (0.12)} &\multirow{2}{*}{3.03(0.04)} & \multirow{2}{*}{3.52 (0.12)}  \\ \cline{3-4}
									      &										                   & \textbf{VAS}                & \textbf{2.24 (0.27)}     & & &                        \\ \cline{2-7}
                                          & VAS                                   & VAS                        & 2.30 (0.26)         	  & 2.82 (0.05) &3.01(0.05) & 3.52 (0.12)	                  \\ \hline
\multirow{3}{*}{\textbf{3rd NN layer }}      & \multirow{2}{*}{\textbf{VAS,OPI}}    & VAS,OPI 	  & 2.34 (0.27)      &\multirow{2}{*}{2.58 (0.19)} &\multirow{2}{*}{3.03(0.04)} & \multirow{2}{*}{3.67 (0.13)}	\\ \cline{3-4}
									      &										                   & \textbf{VAS}               & \textbf{2.18 (0.35)}      &   & &                   \\ \cline{2-7}
                                          & VAS                                   & VAS                  	 & 2.24 (0.25)               &2.65 (0.17)	&3.01(0.05) &	3.67 (0.13)          \\ \hline
\multirow{3}{*}{NN input }    & \multirow{2}{*}{\textbf{VAS,OPI}}     & VAS,OPI		 & 2.41 (0.23)        	  &\multirow{2}{*}{2.48 (0.14)} &\multirow{2}{*}{3.03(0.04)} & \multirow{2}{*}{3.67 (0.13)}  \\ \cline{3-4}
									      &										  & \textbf{VAS}                                 & \textbf{2.22 (0.27)}     &   & &                  \\ \cline{2-7}
                                          & VAS                                  & VAS                         & 2.22 (0.30)         	  &2.53 (0.18) &3.01(0.05) & 3.67 (0.13)         \\ \hline

\end{tabular}
\end{adjustwidth}
\end{table*}

\section{Experiments}
{\bf Features and Evaluation Scores.} 
The model takes as input the 66 AAM facial landmarks of each image frame. Each AAM facial landmark is stored as a (x,y) coordinate, resulting in an input feature vector length of 132 when all coordinates are concatenated. Mean Absolute Error (MAE) is used for evaluation as it is commonly used for ordinal estimation tasks \citep{corf}. Additionally, Intra-Class Correlation ICC(3,1) \citep{Shrout1979} is reported as it is commonly used in behavioral sciences to measure agreement between annotators (in this case, the estimated and true VAS). ICC is also used because it indicates when MAE scores are deceptively low. For example, while MAE provides a good indication of accuracy, it fails to capture the correlation between model estimates and their true values. In this application, if a model were to output the mean VAS (3.25) for all input image sequences, then the resulting MAE would be 2.44, supposedly outperforming some of the baseline models. However, in such a case, ICC would strongly penalize the errors, resulting in an ICC of approximately zero.

{\bf Learning.} Learning is performed using 5-fold cross validation. The data is split into 5 folds such that each fold contains the data for 5 persons. In each trial, 4 folds are used for training while 1 is held out for testing. The VAS estimates from each iteration of testing are saved and, after all 5 iterations, concatenated. The estimates obtained for all 200 sequences are then used to compute the MAE and ICC.

We compare the results of DeepFaceLIFT to a baseline RNN model \citep{mikolov2010recurrent}, comprising a 36-node LSTM layer and one dense layer, and HCRF model \citep{gunawardana2005hidden}, comprising 5 hidden states. We do so since both RNNs and HCRFs are common choices for modeling sequential data. We also construct a NN baseline by taking the mean voting (NN-MV) of per-frame estimations for each sequence.\footnote{We also tested other statistics, such as the median, max, etc., to construct the NN-MV baseline. Mean voting provides the best estimates and so is used as the baseline.}

When using statistics derived from VAS estimates alone, DeepFaceLIFT achieves the lowest MAE (2.18), outperforming the baseline NN-MV (2.48), RNN (3.21), and HCRF (3.38) models. Additionally, when compared to the baseline models, DeepFaceLIFT yields the best MAE and ICC in all settings. The final ICC for DeepFaceLIFT (0.35) improves upon that of the NN-MV (0.19), demonstrating the effectiveness of the GP regression model and feature statistics in the second stage of the model. We note from Table \ref{my-label} that using the personalized features yields better results than using non-personalized features, as improvements are seen not only in DeepFaceLIFT, but also in the RNN and HCRF baselines. Appending the personal features to the third layer of the NN in DeepFaceLIFT achieves a lower MAE than appending them to the input features. This improvement suggests that when personal features are appended to the inputs, their effectiveness may be partially ``washed out" by the larger input feature dimension ($D=132$). A discussion on how each personal feature affects the estimation performance is detailed in Sec.4.1 \label{pf}. Likewise, when OPI is used as a second label in the NN-MV baseline model, the MAE and ICC improve upon those obtained from using the VAS label alone. This suggests that OPI may provide control for the subjective, self-reported pain scores, yielding better estimates of VAS. This also provides evidence that having OPI as an additional task improves the model's estimation, as has been previously shown in other multiple learning tasks \citep{caruana1998multitask, ruder2017overview}.
\subsection{DeepLIFT: Interpretable Learning}
\label{exp}
As discussed earlier, NNs can be cumbersome due to their blackbox nature. There are several approaches available that allow researchers to infer how each input feature contributes (ie. how important the feature is) to the final output \citep{zeiler2014visualizing,sundararajan2016gradients}. However, most of these  methods encounter issues, such as saturation and thresholding, and are computationally expensive. DeepLIFT is robust because it estimates input contributions while remaining computationally efficient and avoiding gradient discontinuities \citep{shrikumar2016not}.

\begin{figure*}
	\centering
	\includegraphics[width=0.7\textwidth]
{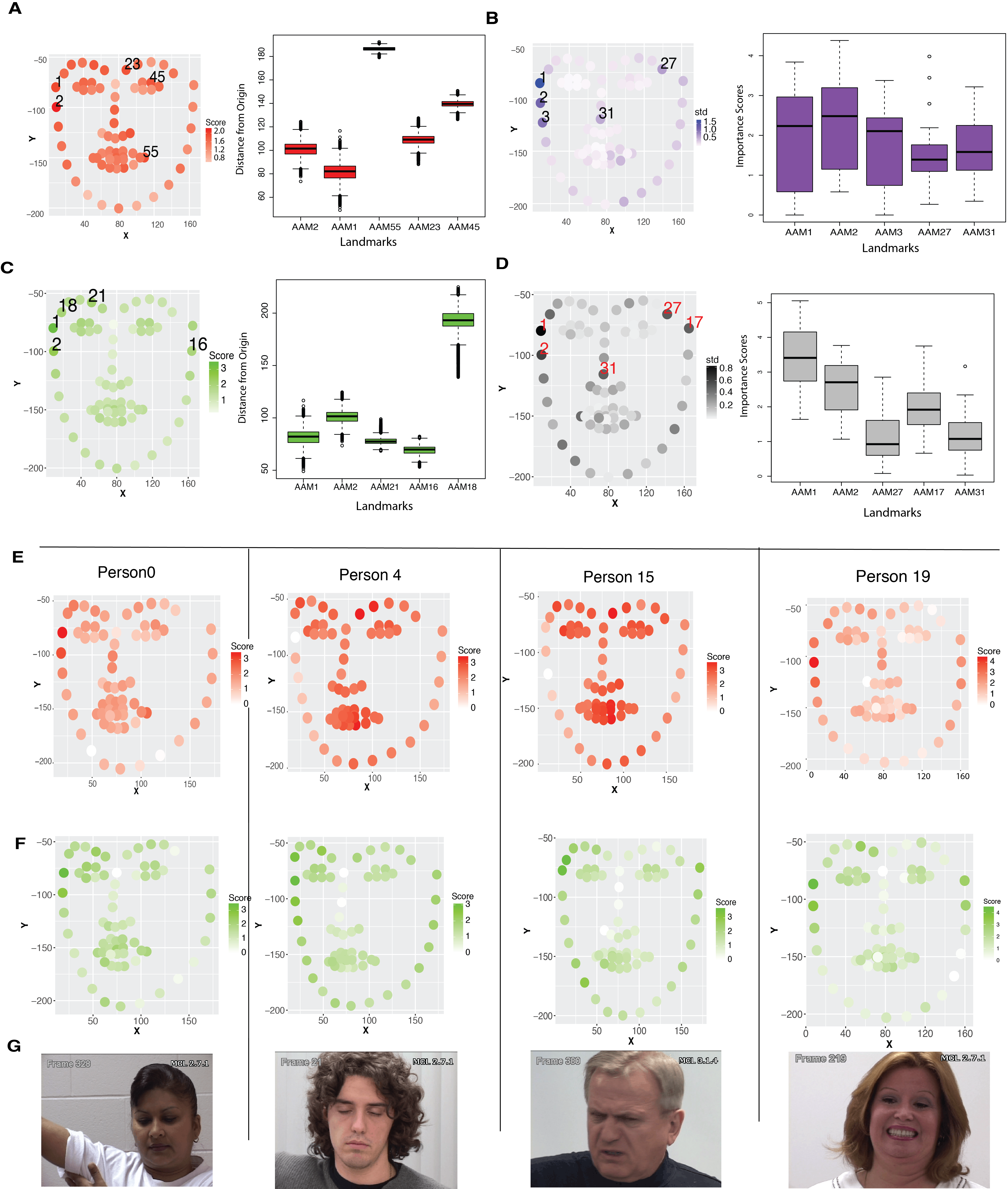}
	\vspace{-0.5em}
    \caption{Contribution score estimates generated by DeepLIFT for each landmark. Facial images illustrate various statistics about the contribution scores, where darker colors indicate more positive values. A) Facial image shows the mean contribution score of each AAM facial landmark for VAS estimation. Boxplot shows distribution of distance from the origin for the 5 facial landmarks with the highest mean contributions to illustrate their movement. B) Facial image shows the standard deviation of contribution scores of each facial landmark for VAS estimation. Boxplot shows distributions for the 5 landmarks with the highest standard deviations. C) Facial image shows the mean contribution score of each facial landmark for OPI estimation. As in A, boxplot shows distribution of distances from the origin for the 5 facial landmarks with the highest mean contributions. D) Facial image shows the standard deviation of contribution scores of each facial landmark for OPI estimation. As in B, boxplot shows distributions of 5 facial landmarks with highest standard deviations. E-G). Four randomly selected subjects from different age, gender, and skin-tone groups. E) The contributions of facial landmarks for VAS estimation, where the contribution scores have been normalized for each subject such that the scores sum to 100. F) Contributions of facial landmarks for OPI estimation. G) Images of the subjects as provided in the dataset.}
    \vspace{-1em}
    \label{fig:DeepLIFT}
\end{figure*}

DeepLIFT estimates the contributions of input features by performing backpropagation through each neuron \citep{shrikumar2016not} then calculating the difference between each neuron's computed activation and its respective ``reference activation" \citep{shrikumar2016not}. These differences are taken between specific input and output activations, allowing DeepLIFT to explain variations in input feature contributions \citep{shrikumar2016not}. 

To clarify, let $V_{s}^{i}=\{v_{s}^{i},\dots,v_{s}^{i}\}$ and $X=\{x_1, \dots,x_T\}$ represent the output and input, respectively. For each subject, features are first normalized using Z-scores. Next, DeepLIFT uses a zero-vector $X_{0}=\{x_{r_1}, \dots,x_{r_{T}}\}$ as a reference activation for the input.  $X_0$ is then passed as an input into the trained NN to obtain the output $f(X_{0})$, which is then used as the reference activation for the output. Each input is assigned a contribution score $C_{\Delta x_{f} \Delta V_{s}^{i}}$ such that $\Delta V_{s}^{i}$=$\sum C_{\Delta x_{f} \Delta V_{s}^{i}}$. For NNs with multiple layers, $\Delta V_{s}^{i}$  is backpropagated to the input layer using the standard chain rule as follows: if $x$ is an input and $h$ is a neuron in a hidden layer, a multiplier that is used to calculate the contribution of $\Delta x$  to $\Delta y$ is defined as: $m_{\Delta x \Delta y} = \frac{C_{\Delta x \Delta h}}{\Delta x}$. Assuming that $y$ is the target output and only $h$ is between $x$ and $y$, then $m_{\Delta x \Delta y} = \sum{m_{\Delta x \Delta h}m_{\Delta h \Delta y} }$.
 
The contribution of each landmark to the frame-level VAS estimation is shown in Fig.\ref{fig:DeepLIFT}A, where darker colors indicate higher contributions  of an input feature to the final output. In this figure, an average face is computed using all image frames to illustrate the locations of each AAM landmark. The coordinates of the top 5 AAM landmarks for the corresponding contribution score statistic are indicated as well. An interesting result is that DeepLIFT finds the 5 highest average input contributions in one subset of facial landmarks when estimating VAS and another subset when estimating OPI, with little overlap. One possible explanation for this difference is that the model may have simply converged on different sets of parameters. Another possibility may come from the different ways in which VAS and OPI scores are obtained. In other words, when obtaining OPI, external observers may perceive a subject's pain differently, focusing on facial regions that differ from the ones engaged by the subject. 
Additionally, Fig. \ref{fig:DeepLIFT}B. shows the standard deviation of contribution scores across subjects, and Fig. \ref{fig:DeepLIFT} C and D show the mean and standard deviation of contribution scores, respectively, for estimating OPI, similarly to Fig. \ref{fig:DeepLIFT}A and B. 

Examples of four different subjects are given in Fig. \ref{fig:DeepLIFT}E, F and G to illustrate the relative importance of using VAS and OPI labels and are accompanied by pictures of the subjects. These results suggest that facial landmarks also contribute differently to VAS and OPI estimation and that their contributions vary among subjects. Based on Fig. \ref{fig:DeepLIFT}, there also appears to be slightly greater contributions from landmarks on the right side of the faces. This may be attributed to the redundancy of AAM landmarks due to facial symmetry or to how the camera is positioned relative to the faces.

\begin{figure}
	\centering
	\includegraphics[width=0.8\textwidth]{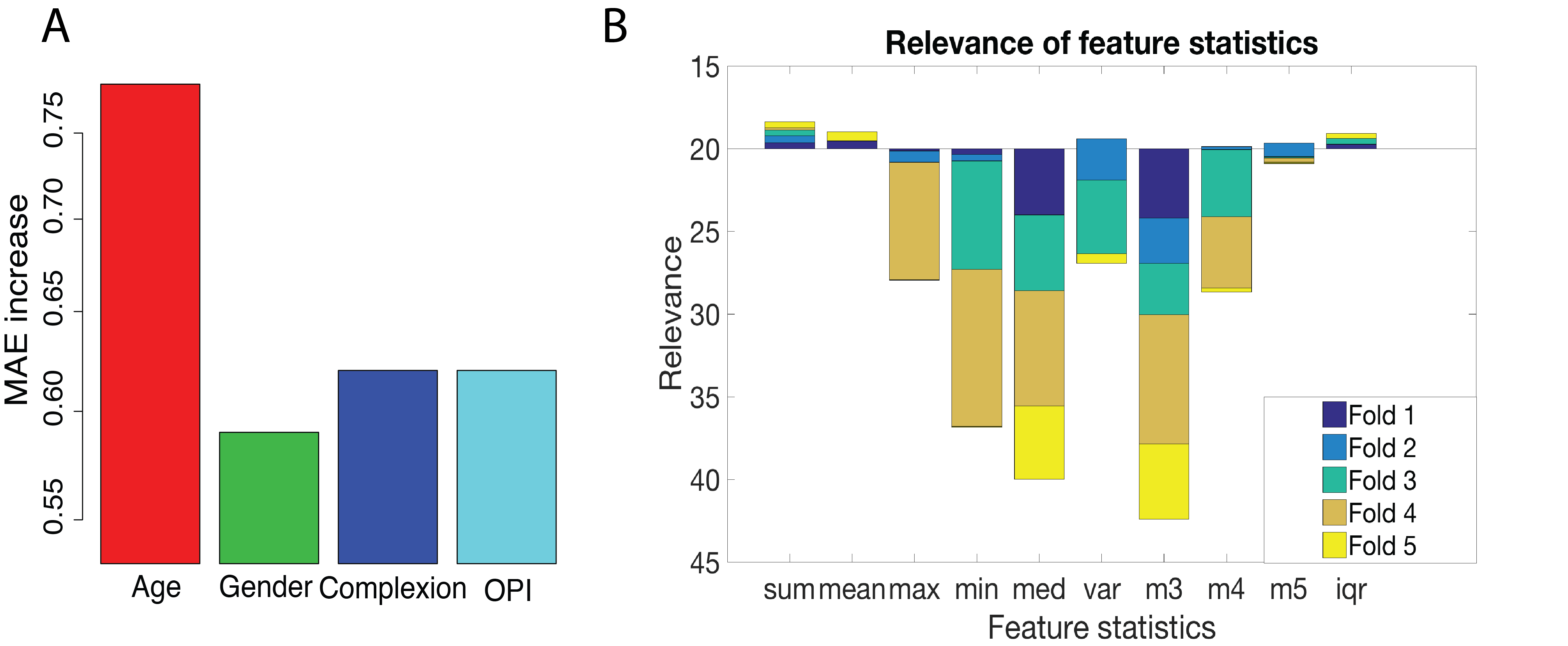}
    \caption{ A) Contributions of personal features to VAS estimation. The Y-axis indicates the MAE increase (decrease in performance) when the specific personal feature is excluded from training. B) The aggregated relevance results for the feature statistics (Sec.\ref{stagetwo}) as computed by the RBF-ARD kernel in the GP model. Lower or more negative values indicate that the statistic contributes more to the final output. (In this figure, -ln($l_i)$ for each feature statistic $i$ is shown). The above figure shows that the sum, mean, and IQR of the frame-level estimates are the most important statistics.}
    \label{fig:PFandWeights}
\end{figure}

{\bf Personal Features.} The importance of each personal feature toward VAS estimation is approximated by excluding each feature individually in the first stage, then calculating the resulting MAE. The importance of including the OPI label in Stage 1 is assessed by including and excluding OPI from from the output, then comparing the resulting MAEs. Of the personal features, age improves VAS estimates the most while all personal features improve final VAS outputs (Fig.\ref{fig:PFandWeights}A). These findings are consistent with clinical studies which indicate that age, gender, and other personal features affect pain perception and expression. Additionally, OPI labels seem to contain information that improves VAS estimates. 

{\bf Statistics.} We aggregated the relevance of each feature statistic using 5-fold cross validation, the results of which are shown in Fig.\ref{fig:PFandWeights}B. For any particular feature statistic, the lower the numerical kernel value is, the more the statistic contributes to the final VAS estimate. According to the model, the 3 statistics that contribute the most are $\Sigma (P), \mu (P),$ and $IQR(P)$, with $\Sigma (P)$ being the most important. This finding suggests that the average of the frame-level estimates and the length of the corresponding image sequence both contribute greatly to the estimation of VAS, as the two are related by $\mu(P) = \frac{\Sigma (P)}{L}$, where $L$ is the length of the sequence. The connection between the length of an image sequence and the highest contributing statistics is indicative of a correlation between the length of the captured pain-causing experience and the resulting VAS score.\\

\section{Discussion and Conclusion}
We presented a two-stage interpretable machine learning model  that uses face images to automatically estimate subjective, self-reported VAS pain scores. DeepFaceLIFT demonstrates that a weakly supervised learning model is able to accurately estimate VAS from image sequences, and that using personalized features and OPI scores during training improves model performance. Applying a GP model to estimate sequence-level VAS scores using statistics derived from frame-level embeddings further improves model performance. 

An additional contribution of DeepFaceLIFT is its ability to track the contribution of features in both stages of the learning. This makes it possible to explore the relationships between input features and output estimates, providing greater insight into how facial expressions influence pain perception and estimation. The contributions of input features estimated using DeepLIFT suggest that experienced and observed pain are perceived differently and that the contributions from specific facial landmarks towards pain detection varies considerably among subjects. More analysis, however, is needed to verify whether these differences reflect the model's behavior alone or are also a result of the differences in how subjects (VAS) and observers (OPI) rate pain.

Finally, we identified and examined important personal features and statistical embeddings, elaborating on how both improve model performance. We also showed that including personal features at different locations in the first stage of DeepFaceLIFT results in different performance gains, where optimal performance is achieved by embedding them at an intermediate layer of the NN. Furthermore, by exploiting the benefits of multi-task learning, we obtain a more robust estimation of VAS \citep{caruana1998multitask}. We attribute this to the fact that OPI comes from an external observer and thus acts as a sort of grounding for estimating subjective VAS scores.

To summarize, we proposed a personalized machine learning approach for estimating VAS. We illustrated the personalization process in several different settings using NN and GP-based model, the results of which indicate the benefits of the feature-level personalization. In the future, we plan to further investigate the relationships between VAS and other pain scores (such as PSPI) as well as how they relate to facial AUs. As context is another important aspect not explored in this paper, we plan to derive more advanced statistics that could potentially capture additional information and improve estimates of subjective pain. Finally, we hope that these findings will advance applications of pain estimation in clinical settings. 

\section{Acknowledgements}
The work of O. Rudovic is funded by the European Union H2020, Marie Curie Action - Individual Fellowship no. 701236 (EngageMe). The work of D. Liu is funded by the Vest Scholarship and Wellcome Trust Fellowship.  We also thank Daniel Lopez Martinez (Media Lab, MIT) and Kelly N Peterson (MIT) for their help in the early stages of the project.  


\small
\bibliography{Mendeley,MendeleyOggie,library,bibfile,ijcai17}

\end{document}